%% file: bmvc_review.tex
\documentclass{bmvc2k}
\usepackage{bm, multirow, amsmath, amssymb, array, floatrow} 

\title{A CNN Based Approach for the Near-Field Photometric Stereo Problem}

\addauthor{Fotios Logothetis}{flogothetis@crl.toshiba.co.uk}{1}
\addauthor{Ignas Budvytis}{ib255@cam.ac.uk}{2}
\addauthor{Roberto Mecca}{rmecca@crl.toshiba.co.uk}{1}
\addauthor{Roberto Cipolla}{rc10001@cam.ac.uk}{2}
\addinstitution{
Cambridge Research Laboratory \\
Toshiba Europe Ltd.\\
Cambridge, UK
}
\addinstitution{
 University of Cambridge\\
 Cambridge, UK
}

\runninghead{LOGOTHETIS ET AL.}{A CNN Based Approach for the NF-PS Problem}

\def\etal{\emph{et al}\bmvaOneDot}
\usepackage{amsmath,amsfonts,amssymb,amsthm}
\usepackage{newpxtext} 
\usepackage{hyperref}
\usepackage[hang,flushmargin]{footmisc}
\usepackage{placeins}
\begin{document}

\maketitle
\vspace{-0.15cm}
\begin{abstract}
Reconstructing the 3D shape of an object using several images under different light sources is a very challenging task, especially when realistic assumptions such as light propagation and attenuation, perspective viewing geometry and specular light reflection are considered. Many of works tackling Photometric Stereo (PS) problems often relax most of the aforementioned assumptions. Especially they ignore specular reflection and global illumination effects. In this work, we propose the first CNN based approach capable of handling these realistic assumptions in Photometric Stereo. We leverage recent improvements of deep neural networks for far-field Photometric Stereo and adapt them to near field setup. We achieve this by employing an iterative procedure for shape estimation which has two main steps. Firstly we train a per-pixel CNN to predict surface normals from reflectance samples. Secondly, we compute the depth by integrating the normal field in order to iteratively estimate light directions and attenuation which is used to compensate the input images to compute reflectance samples for the next iteration. To the best of our knowledge this is the first near-field framework which is able to accurately predict 3D shape from highly specular objects. Our method outperforms competing state-of-the-art near-field Photometric Stereo approaches on both synthetic and real experiments.

\end{abstract}

\input{sections/Introduction.tex}

\input{sections/RelatedWorks.tex}


\input{sections/Method.tex}

\input{sections/ExperimentalSetup.tex}

\input{sections/Experiments.tex}

\input{sections/Conclusion.tex}

\FloatBarrier
\bibliography{egbib}
\end{document}

%% file: sections/Introduction.tex
\vspace{-0.01cm}
\section{Introduction}
\label{sec:introduction}
\vspace{-0.01cm}
\begin{figure}[t]
\centering
\includegraphics[width=0.95\columnwidth,trim={0.5cm 0.3cm 0cm  0.2cm},clip]{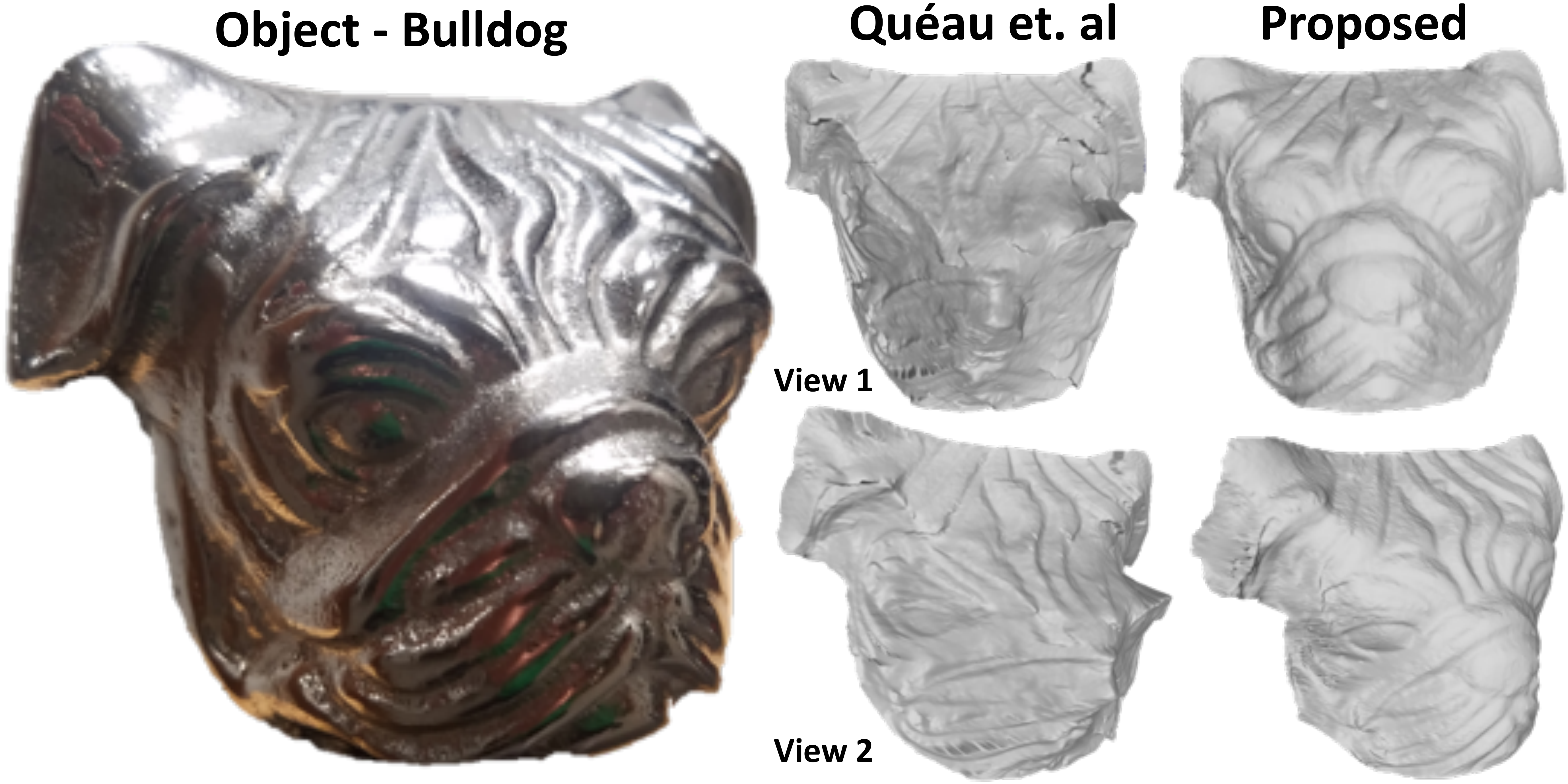}
\vspace{-0.25cm}
\caption{Our proposed approach accurately reconstructs highly specular objects, clearly outperforming the previous state-of-the-art~\cite{queau2018led}.}
\label{fig:intro}
\end{figure}

Retrieving the 3D shape of an object from observations under varying illumination is a very challenging problem in Computer Vision introduced by \cite{Woodham1980} under the name of Photometric Stereo (PS). The mathematical model of the Photometric Stereo problem proposed in this work relied on four main assumptions: orthographic viewing geometry, diffuse light reflection, uniform light propagation and the lack of ambient nuisance. 
Due to restrictive assumptions, the method of \cite{Woodham1980}  was limited to very narrowly specified scenarios. Since then an extensive research has been carried out to relax these assumptions. 

With the aim of solving the PS problem under more realistic conditions, researchers have modelled perspective viewing geometry \cite{Prados2003, Tankus2005,Onn1990}, specular light reflections \cite{MeccaQLC2016} 
and point light sources parameterising  radial propagation of light \cite{Iwahori1990point,Clark1992}. These effects lead to highly non-linear models requiring sophisticated optimisation strategies \cite{WMBK14,queau2018led}. As the complexity of the models becomes intractable, especially when dealing with realistic material properties, a lot of recent works have opted to neglect specular highlights and instead rely on robust optimisation techniques \cite{Ikehata2012Robust,6909677}. Furthermore, materials that reflect light in a non-diffuse manner are prone to a number of complex physical effects which make the explicit mathematical modeling very hard to invert. 

In fact, these global illumination effects (cast shadows, self reflections, ambient light) are one of the most challenging aspects of PS. \cite{logothetis2016near,YuilleSEB99} tackle the case of fixed ambient light which however is too simple of a model to cover realistic inter-reflections.
The global illumination issue  is firstly adequately addressed in \cite{ikehata2018cnn} by employing a Convolutional Neural Network (CNN). 
This method works by arranging reflectance (BRDF) samples into a fixed size observation map for each pixel.  Observational maps are then provided as an input to the CNN which is trained to output a surface normal per pixel. \cite{logothetis2020pxnet} extends this work and shows how a training data augmentation strategy can be used to deal with general reflectance BRDF such as MERL dataset~\cite{Matusik03} or Disney BSDF~\cite{burley2012physically} and global illumination effects in the far-field setting. However, these approaches are only directly applicable to the far-field photometric stereo since the nonlinear light attenuation of near-field images does not allow to directly compute valid observation maps.  

We propose to overcome the aforementioned limitations by using a three step process. Firstly, the effect of the light attenuation is compensated using an estimate of the object geometry, to produce the equivalent far-field reflectance samples. Secondly, a CNN is used to regress pixel normals from these samples. Finally, a numerical integration is used to update the estimate of the object geometry for the next iteration step. In addition, since during test time the observation map sample space is highly limited due to a fixed LED capture setup, we render our training observational maps from a virtual copy of this near-field image capture device. In this way we significantly reduce the training data variation and allow ourselves for various geometric augmentations such as pixel depth variation. 

We evaluate our method on both artificial and real near-field image datasets. We significantly outperform competing approaches~\cite{queau2018led,logothetis2017semi} on both datasets ,improving by more than $6^o$ on the average angular error of the synthetic one.


The rest of this work is divided as follows. Section 2 discusses relevant work in Photometric Stereo. Section 3 provides details of our proposed method. Sections 4 and 5 describe the experiment setup and corresponding results.


%% file: sections/RelatedWorks.tex
\vspace{-0.2cm}
\section{Related Work}
\label{sec:relatedWorks}
\vspace{-0.1cm}
In this section we provide an overview of the relevant latest improvements. 
For a detailed, fairly recent PS survey, refer to \cite{AckermannG15}. 

\noindent
\textbf{Near-Field PS.} Differently from the classical far-field PS, near-field approaches assume that the illumination is non-linear with respect to the position of the light sources, thus making analytical models a lot more complicated and harder to solve in practice. 
Some works embedded these non-linearities in a PDE-based formulation so the depth is directly calculated without passing through the computation of the normal field \cite{Mecca2014near}. 
\cite{Lee1991glaucoma,SmithFang2016} took advantage of image ratios in order to eliminate the dependence on the surface albedo and thus reduce the number of unknowns. Image ratios were also used in the variational framework of \cite{MeccaQLC2016} in order to make the approach more robust to specular highlights by unifying diffuse and Blinn-Phong specular \cite{Blinn:1977} reflections into a single mathematical formulation. This general variational framework is also applicable in a weekly calibrated setting \cite{logothetis2017semi} or even a volumetric one \cite{logothetis2019differential}. 
Recently, the LED-based approach introduced by Qu\'eau \etal \cite{queau2018led} presented a convoluted variational approach based on alternating weighted least-square scheme also capable of calibrating the light brightness of the light sources. Finally, \cite{liu2018near} exploited a circular LED setup to compute the relative mean distance between the camera and the object.

\noindent
\textbf{Deep Learning based approaches for PS.} Computer graphics is a well understood topic and many tools capable of rendering highly non-linear irradiance equations are publicly available\footnote{\url{www.blender.org} and \url{www.disneyanimation.com/technology/brdf.html}} \cite{Matusik03}.
This allowed to create reliable datasets for supervised DL approaches. The potential of DL for solving the PS problem can be divided in two main advantages. Firstly, CNNs have the capability of inverting highly non-linear reflectance models comprising of numerous physically based parameters. Secondly, CNNs can be made to deal with the complicated real world imperfection (shadows, self reflections, noise) through the use of data augmentation.
So far, several DL approaches have been proposed. 
A preliminary work by \cite{TangSH12a, Hinton09} 
considered diffuse reflection only. \cite{YuS17} 
proposed a library where set of novel layers can be incorporated into a generic neural network to embed explicit models of photometric image formation. More recently, several approaches have tackled the problem of reconstructing complex objects. \cite{santodeep} proposed a method to find correspondences between simulated observation rendered by the MERL BRDF dataset \cite{Matusik03} and the normal map of the target object,  handling non-local effects using a dropout strategy.
\cite{JuQZDL18} leveraged DL to learn the information from multispectral images to get RGB information. 
\cite{taniai2018neural} proposed generating training data on the go to minimise the image re-projection error. Although this method is a training data free approach, the whole procedure is slow. 
Recently, \cite{ChenHW18} proposed rendering patches of different surface materials in order to get training data. This method is also extended in \cite{chen2019self} for solving the uncalibrated PS. 
 Finally, \cite{ikehata2018cnn} proposed arranging all the reflectance samples of a pixel (i.e. different illumination images in the far-field setting) into a fixed size \textit{observation} map which is used by a CNN to regress pixel normals. The CNN is essentially learning to invert the BRDF with added robustness to global illumination effects, as training data are made with physics based renderer. In  \cite{logothetis2020pxnet}, this method was extended by simplifying the training procedure. 
However, none of these DL approaches is directly applicable to the NFPS problem and non-linear attenuation from point light sources together with the viewing direction dependency drastically increase the problem space exploding the training data requirements. To address this issue, the NF problem is re-arranged appropriately into a far-field one allowing the use of the observation maps. Finally, to maximise performance, only the appropriate subset of maps is sampled at train time.




%% file: sections/Method.tex
\begin{figure}[t]
\centering
\includegraphics[width=0.995\columnwidth]{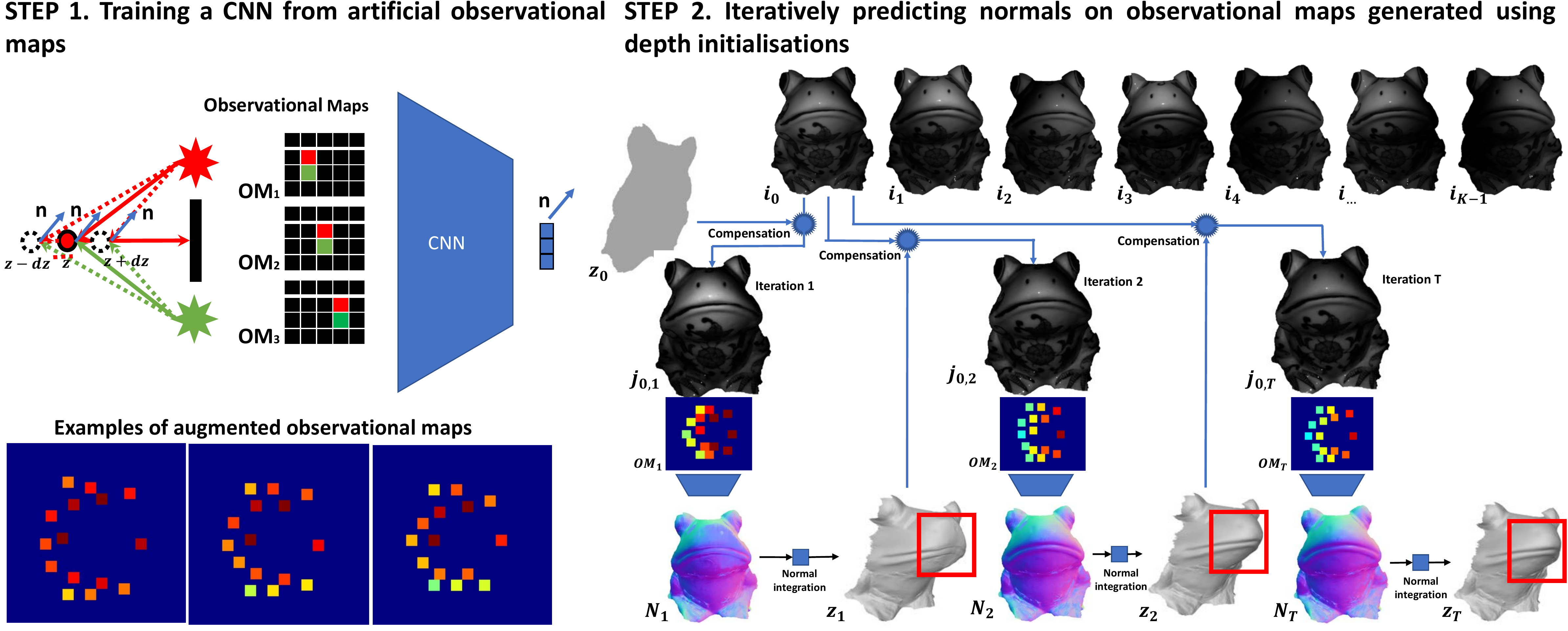}
\vspace{-0.25cm}
\caption{This figure illustrates two key steps of our proposed approach. On the left, the network training is illustrated consisting of sampling points inside the camera's frustrum and rendering the respective observation maps. As the depth will only be approximately known at test time, this is slightly perturbed before mapping resulting to a structured change of the map (shown at bottom left: middle is actual depth (10~cm), left and right correspond to $\pm1$cm respectively. On the right, the reconstruction process is shown. Images $i_{0}\cdots i_{K-1}$ are used with conjuction with previous depth estimate to compensate for light attenuation ($j_{0}\cdots j_{K-1}$), compute observation maps (shown for pixel at image center here), regress normals and finally update the shape. Note the improvement of the shape of frogs beak (red square) from iteration 1 to iteration 2.} 
\label{fig:method}
\end{figure}

\clearpage
\vspace{-0.02cm}
\section{Method}
\vspace{-0.01cm}
\label{sec:method}

In this section we describe our method for tackling the near field Photometric Stereo problem. In particular, we provide both the details of the assumed image formation model and how normals can be predicted for near-field images by using CNN's trained on far-field reflectance samples (also see Figure~\ref{fig:method}) . 



\noindent
\textbf{Near  field Modeling.}
Similar to \cite{Mecca2014near}, we assume calibrated point light sources at positions $\mathbf{P}_m$ (w.r.t the camera center at $\mathbf{0}$) resulting in variable lighting vectors  $\mathbf{L}_m=\mathbf{P}_m-\mathbf{X}$. Here $\mathbf{X}=[x,y,z]^ \intercal$ is the 3D surface point coordinates. 
We also model the light attenuation considering the following non-linear radial model of dissipation
\begin{equation}
\label{eq:attenuation}
a_m(\mathbf{X})=\phi _m \frac{( \hat{\mathbf{L}}_m (\mathbf{X}) \cdot \hat{\mathbf{S}}_m)^{\mu _m}}{||\mathbf{L}_m (\mathbf{X})||^2},
\end{equation}
where $\hat{\mathbf{L}}_m=\frac{\mathbf{L}_m}{||\mathbf{L}_m||}$ is the lighting direction, $\phi _m$ is the intrinsic brightness of the light source,~$\hat{\mathbf{S}}_m$ is the principal direction (i.e. the orientation of the LED point light source) and $\mu _m$ is an angular dissipation factor. Defining $\hat{\mathbf{V}}=-\frac{\mathbf{X}}{||\mathbf{X}||}$ as the viewing vector, the general image irradiance equation becomes: 
\begin{equation}
\label{eq:irad}
i_m= a_m \text{B}(\mathbf{N},\hat{\mathbf{L}}_m,\hat{\mathbf{V}},\rho)
\end{equation}
Here $\mathbf{N}$ is the surface normal. B is assumed to be a general BRDF and $\rho$ is the surface albedo (allowing for the most general case, images and $\rho$ are RGB and the reflectance is different per channel). In addition, we allow for the possibility of global illumination effects (shadows, self reflections) which are incorporated into B. This can be re-arranged into a BRDF inversion problem as:
\begin{equation}
\label{eq:brdf_inv}
j_m=\frac{i_m}{ a_m}= \text{B}(\mathbf{N},\hat{\mathbf{L}}_m,\hat{\mathbf{V}},\rho).
\end{equation}

\noindent
We note that $\hat{\mathbf{V}}$ is known but $\mathbf{L}_m$ and $a_m$ are unknowns due to the nonlinear dependence on $z$. Our objective is to recover the surface normals $\mathbf{N}$ and depth $z$ 

\noindent
\textbf{Normal Prediction.} The first step of our method includes training a CNN for per-pixel normal prediction using BRDF samples. This is done through the the observational map parameterisation introduced by \cite{ikehata2018cnn} in order to tackle the far-field photometric stereo problem. Note that this is equivalent to BRDF inversion under the special case of  $\hat{\mathbf{V}}=[0,0,1]$. 
As described in~\cite{ikehata2018cnn} an observational map records relative pixel intensities (BRDF samples) on a 2D grid (e.g. $32 \times 32$) of discretised light directions. Such a representation is highly convenient for use with classical CNN architectures as it provides a 2D input and is of fixed shape despite a potentially varying number of lights. While~\cite{ikehata2018cnn} proposes to train CNNs on rendered images of objects, it is shown in~\cite{logothetis2020pxnet} that simpler per-pixel renderers can be used instead, making the training procedure much faster and simpler. We use the latter approach in this work. In addition, the known viewing direction can be incorporated by augmenting the map with another two channels, set to $\hat{\mathbf{V}}_x$ and $\hat{\mathbf{V}}_y$ respectively. 

\noindent
\textbf{Adapting to the near-field setup.} In order to solve the near-field PS problem for a specific capture setup, we adapt the training procedure to only sample observation maps which are plausible at test time. We note that our capture hardware has a fixed number of 15 lights rigidly attached to a circuit board around the camera looking at objects at a limited depth range of around 10-20~cm\footnote{When objects are too close to the camera, their images are out of focus; while when being too far, surface details cannot be resolved.}. 
Therefore, instead of sampling a random set of light directions as in \cite{logothetis2020pxnet}, we sample 3D points inside the camera frustrum. Surface normal and material parameters are sampled independently from the point position so as to be able to deal with general surfaces. After the point parameters are sampled, the 15 images are rendered using the renderer of \cite{logothetis2020pxnet} with the addition of near-field light attenuation (Eq.~\ref{eq:attenuation}). Additional global illumination augmentations for shadows, reflections and ambient light are also applied as in~\cite{logothetis2020pxnet}. In order to get robustness to imprecise depth initialisation at test time, the training procedure involves perturbing the ground truth depth value $z$  by $\delta z\sim \mathcal{N}(0,4mm)$\footnote{The Gaussian distribution encourages the network to be more accurate when $\delta z$ is small and thus get an improvement in the iterative setting.} so that the light directions $\mathbf{\hat{L}}$ (determining the sparsity pattern of the map) and attenuation (determining the values-Eq.~\ref{eq:brdf_inv}) are slightly perturbed. See Step 1 in Figure~\ref{fig:method} for an illustration.



\begin{figure}[t]
\centering
\includegraphics[width=0.995\columnwidth]{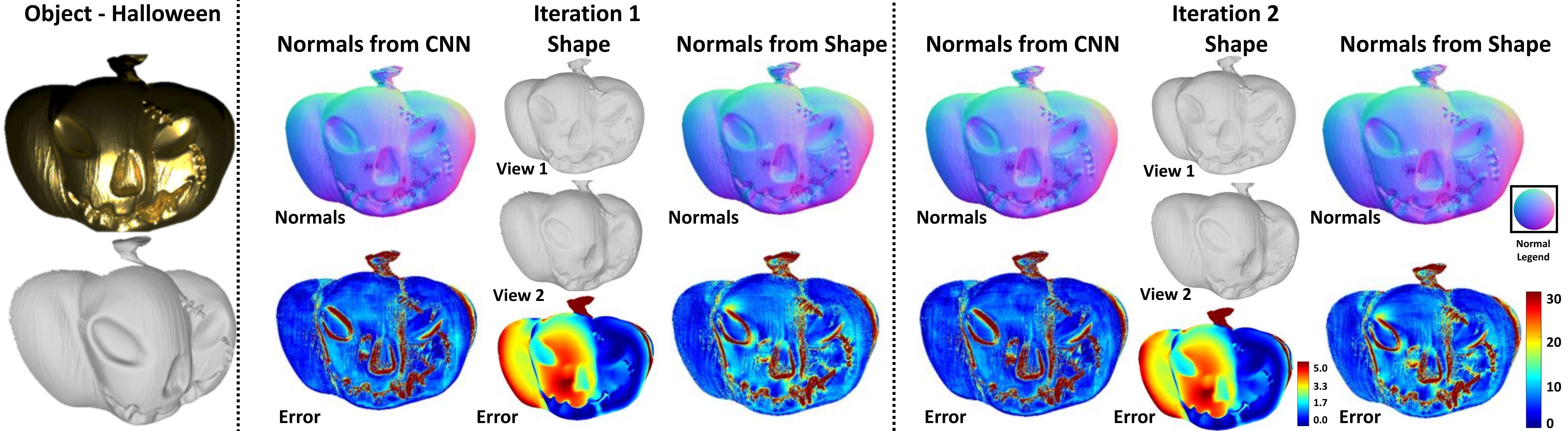}
\vspace{-0.25cm}
\caption{Iterative refinement of the geometry for the Halloween synthetic object. On the left, a sample image and GT shape are shown. The other 2 sections show 2 steps of the iterative refinement with the respective normals (both raw network predictions and differentiated ones), normal error maps and depth error maps. As the difference is minimal between steps 1 and 2, the process is converged.}
\label{fig:normal_pred}
\end{figure}
\noindent
\textbf{Iterative refinement of depth and normals.} Assuming an estimate of normals, the depth can be estimated with numerical integration. This is performed using the $\ell _1$ method of \cite{queau2015edge}. The variational optimisation includes a Tikhonov regulariser $z=z_0$ (weight $\lambda =10^{-6}$) and is solved in a ADMM scheme\footnote{Code ported from \url{https://github.com/yqueau/normal_integration}.}.
As the BRDF samples $j$ (see Equation~\ref{eq:brdf_inv}) depend on the unknown depth, they cannot be directly computed to be input to the network. To overcome this issue, we employ an iterative scheme were the previous estimate of the geometry is used. The procedure involves computing the near to far conversion as described, obtaining a new normal map estimate through the CNN and finally numerical integration. See  Figure~\ref{fig:normal_pred} for an example of intermediate results of our iterative procedure.As it is the case in competing classical methods \cite{logothetis2017semi,queau2018led}, this iterative procedure is initialised with a flat plane at the approximate mean distance.  


%% file: sections/ExperimentalSetup.tex
\section{Experimental Setup}
\label{sec:experimental_setup}
In this section we provide various experimental setup details related to CNN training and datasets used for evaluation.

\begin{figure}[!t]
\subfigure[]{\label{fig:a:roc}
\begin{minipage}{0.34\textwidth}
\includegraphics[width=0.99\textwidth,trim={0.8cm 0.5cm 0.8cm 0.2cm},clip]{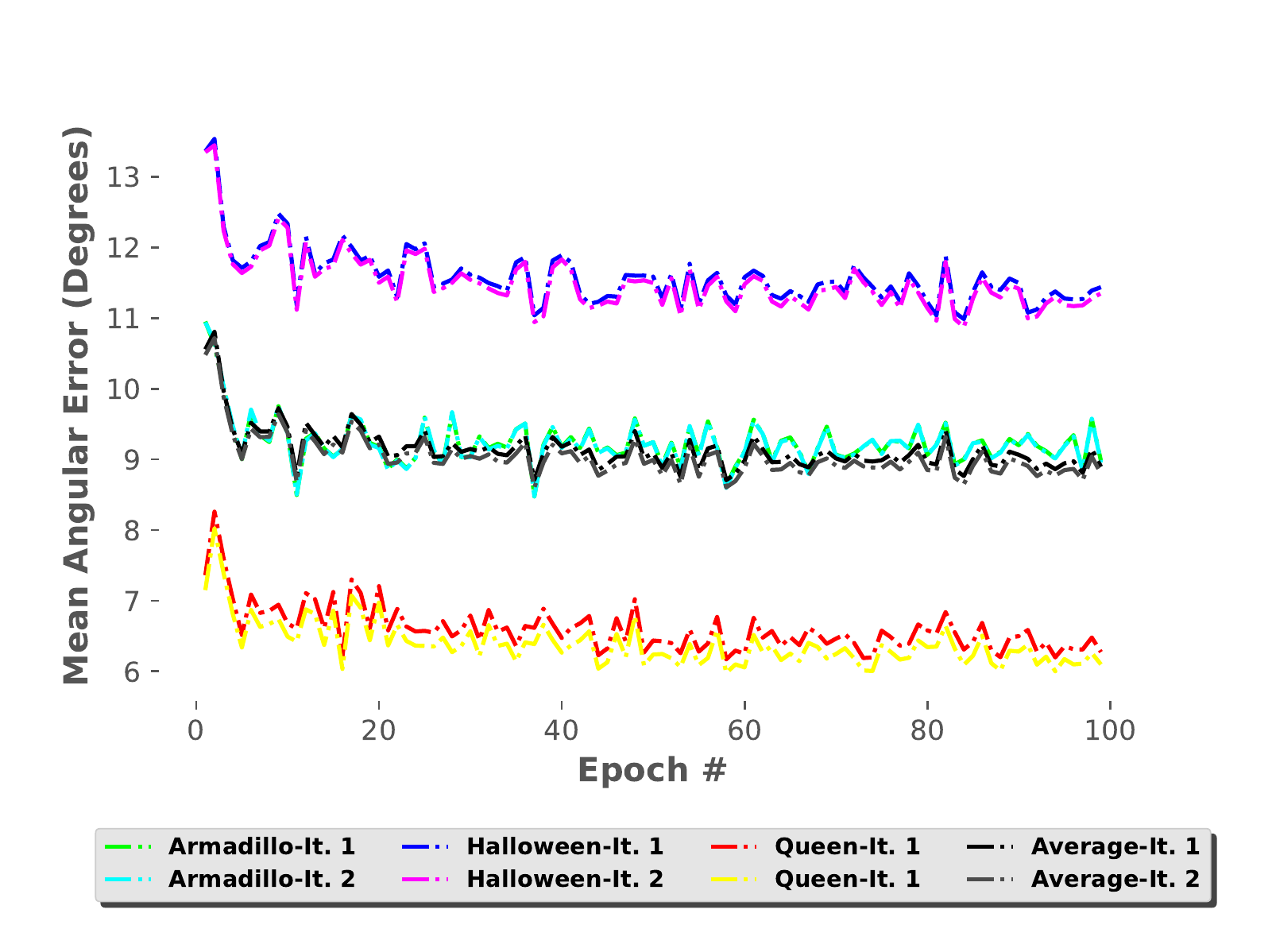}
\end{minipage}
}
\subfigure[]{\label{fig:b:table}
\begin{minipage}{0.63\textwidth}
\centering
\scalebox{0.52}{
\begin{tabular}{ | l | c c c |l|l| }
 \hline
  Method & Armadillo & Halloween & Queen & AVG Norm & AVG Depth  \\ \hline 
GT normals - NfS&5.62&3.5&4.75&\textbf{4.62}& \textbf{2.04}\\ \hline
Naive - CNN-PS~\cite{ikehata2018cnn} - NfCNN&17.25 &17.37 &20.62 &18.41 & - \\
Naive - CNN-PS~\cite{ikehata2018cnn} - NfS&18.15&16.39&21.16&18.57&4.29\\ \hline
Adapted - CNN-PS~\cite{ikehata2018cnn} - NfCNN&11.2&19.06&10.61&13.62&-\\
Adapted - CNN-PS~\cite{ikehata2018cnn} - NfS&12.37&17.85&11.74&13.99&3.04\\
Adapted - PX-Net~\cite{logothetis2020pxnet} - NfCNN&13.11&18.66&7.42&13.06&-\\
Adapted - PX-Net~\cite{logothetis2020pxnet} - NfS&13.87&18.03&8.5&13.47& \textbf{2.10}\\ \hline
Proposed - GT Depth - NfCNN&9.03&11.33&6.31&8.89&-\\ \hline
Proposed - Iteration 1 - NfCNN&9.05&11.47&6.58&9.03&-\\
Proposed - Iteration 1 - NfS&10.82&11.8&8.16&10.26&2.65\\
Proposed - Iteration 2 - NfCNN&\textbf{9.04}&\textbf{11.38}&6.39&\textbf{8.94}&-\\
Proposed - Iteration 2 - NfS&\textbf{10.81}&\textbf{11.7}&8.02&\textbf{10.18}& \textbf{2.63}\\ \hline
Qu{\'e}au et. al \cite{queau2018led} &15.919&24.85&8.68&16.48&2.98\\
Logothetis et. al \cite{logothetis2017semi} &16.07&28.53&\textbf{6.32}&16.97&2.75\\
\hline
\end{tabular}
}
\end{minipage}}
\vspace{-0.25cm}
\caption{(\textit{a}) MAE evolution (during training) curves illustrate the performance of our network (PX-NET) on predicting normals (NfCNN) at first and second iteration . (\textit{b}) Full quantitative comparison on synthetic data. For out method, we report raw normal prediction error as NfCNN and numerically differentiated normals as NfS. Also report NfCNN error when the GT depth is used as initialisation and also NfS error for integrating the GT normals. We compare against state of the art NF methods \cite{queau2018led}, \cite{logothetis2017semi} and naive usage of far field method of \cite{ikehata2018cnn}.}
\end{figure}

\noindent
\textbf{CNN training.} We use the exact architecture of PX-NET \cite{logothetis2020pxnet} which is heavily inspired by CNN-PS \cite{ikehata2018cnn}. 
More specifically, the network includes 7 convolutional, 4 dropout (20\%) and 3 fully connected layers as well as a special log layer. See \cite{logothetis2020pxnet} for more detail. The total number of parameters is $\sim4.5$ million. The network was implemented in Keras of Tensorow 2.0 The training batch size was set at 1600 with 10000 batches per epoch (16 million maps). We trained for 100 epochs which took around one day on 3x  NVIDIA GeForce 1080. 
We trained the network using the mean squared error loss function on the normals. The Adam optimiser with basic LR of $10^{-3}$ and LR decay of $0.1\%$ per epoch after epoch 10 was used. The model test performance evolution over time (epochs) can be seen in Figure~\ref{fig:a:roc}. The reconstruction computation time was 2-3 minutes with the bottleneck being the python implementation of the numerical integration.

\noindent
\textbf{Datasets.} We evaluate our method on real data captured with a custom made setup. This consists of a printed circuit board with 15 white bright LEDs rigidly attached to a camera FLEA3 3.2 MegaPixel with a 8mm lens. The LEDs are evenly spaced around the camera at a maximum distance of 6.5cm and are placed to be co-planar with the image plane. We captured four image sequences namely a metallic-silver Bulldog statue (Figure~\ref{fig:intro}), a metallic-gold Bell, a porcelain Frog as well as a mutli-object scene featuring a shiny wooden elephant statue in front of a porcelain Squirrel (Figure~\ref{fig:main_real_qualitative}). The objects are placed around 15cm away from the camera and the initial depth estimate is approximately measured with a ruler. As there is no ground truth available for these objects, the evaluation on this dataset is only qualitative. Quantitative evaluation is performed using a synthetic dataset rendered with Blender. This artificial light source configuration 
closely mimics our real capture setup. 
The Cycles render engine with the Disney BRDF is used to generate realistic global illumination effects such as cast shadows and self reflections. We present three objects here namely Queen, which is has a dielectric specular material, Halloween is purely metallic and Armadillo which has an intermediate material. The materials are chosen to match the far-field dataset of \cite{logothetis2020pxnet}.

\noindent
\textbf{Evaluation Protocol.} We compare our method against the far-field CNN approaches of \cite{ikehata2018cnn} and \cite{logothetis2020pxnet}, and the near field methods of \cite{logothetis2017semi}\footnote{Code from \url{https://github.com/fotlogo/semi_calib_ps_cvpr2017}}  and \cite{queau2018led}\footnote{Code from \url{https://github.com/yqueau/near_ps}} (disabling the light source calibration parameter). For the synthetic experiments, the evaluation metric is the mean angular error (MAE) of the normal map (degrees) as well as the average depth error (mm). We note that the depth error is less meaningful as all these methods assume a continuous surfaces and thus a systematic error is introduced in the occlusion boundaries.

%% file: sections/Experiments.tex
\section{Experiments}
\label{sec:experiments}

In this section we present various experiments on the synthetic and real datasets introduced in the previews section. These include an ablation study of the numerical integration method,  a naive usage of far-field state-of-the-art networks as well as qualitative and quantitative comparison with NF competitors. 

\begin{figure}[t]
\centering
\includegraphics[height=6.0cm,width=0.995\columnwidth]{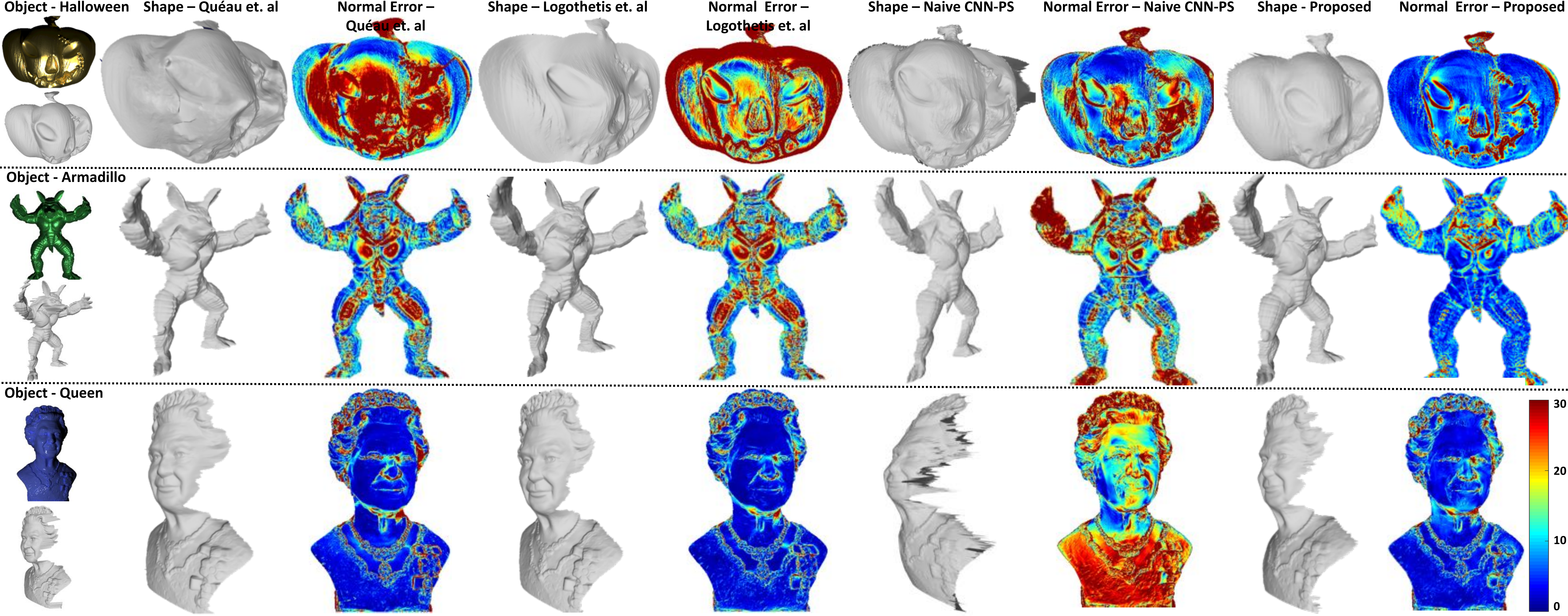}
\vspace{-0.25cm}
\caption{Visualisation of the results of Figure~\ref{fig:b:table}. Optimisation based methods of Logothetis et. al~\cite{logothetis2017semi} and Qu{\'e}au et. al~\cite{queau2018led} fail on the reflectance of the metallic Halloween and so they exhibit very high MAE. In contrast, the dielectric specular Queen is more consistent with their assumed reflectance model and thus exhibit smaller MAE. The naive far-field method CNN-PS~\cite{ikehata2018cnn} also has significant global deformation as it does not model the light attenuation effect. The proposed approach is able to cope with all, highly different, materials and achieves the state-of-the-art performance.}
\label{fig:main_synthetic_qualitative}
\end{figure}
\clearpage
\noindent
\textbf{Shape integration.} The first experiment we conducted aimed at calibrating the quality of the numerical integration of the normal map. As no realistic depth map is C2 continuous, GT normals are not compatible with the GT depth. Indeed, integrating the GT normals and then re-calculating them with numerical differentiation introduces $4.62^o$ (Figure \ref{fig:b:table} top) MAE on average which is by no means negligible. Therefore, for the rest of the experiments we present two figures namely MAE on the raw network predictions (NfCNN) and MAE after differentiation of the surface (NfS). Since the network is trained to regress normals, the first figure tends to be slightly lower.

\noindent
\textbf{Naive usage of far-field networks.} The next experiment consists of naively using the far-field methods \cite{ikehata2018cnn} \cite{logothetis2020pxnet} with near field images without any attenuation compensation and by using the average light direction for each LED. The predicted normals are also integrated using our method in order to have a qualitative shape comparison (Figure~\ref{fig:main_synthetic_qualitative}). As expected, the networks are able to cope with the complicated reflectance and global illumination effects (no bumps at specular highlights); however, not accounting for the variable lighting direction and attenuation results to severe distortion, which is also depicted in larger MAE. To add to this point, using these network as part of our iterative process drastically improves the result and in fact they outperform the classical optimisation methods of \cite{logothetis2017semi} and \cite{queau2018led}. See Table~\ref{fig:b:table}.

\begin{figure}[t]
\centering
\includegraphics[width=0.995\columnwidth]{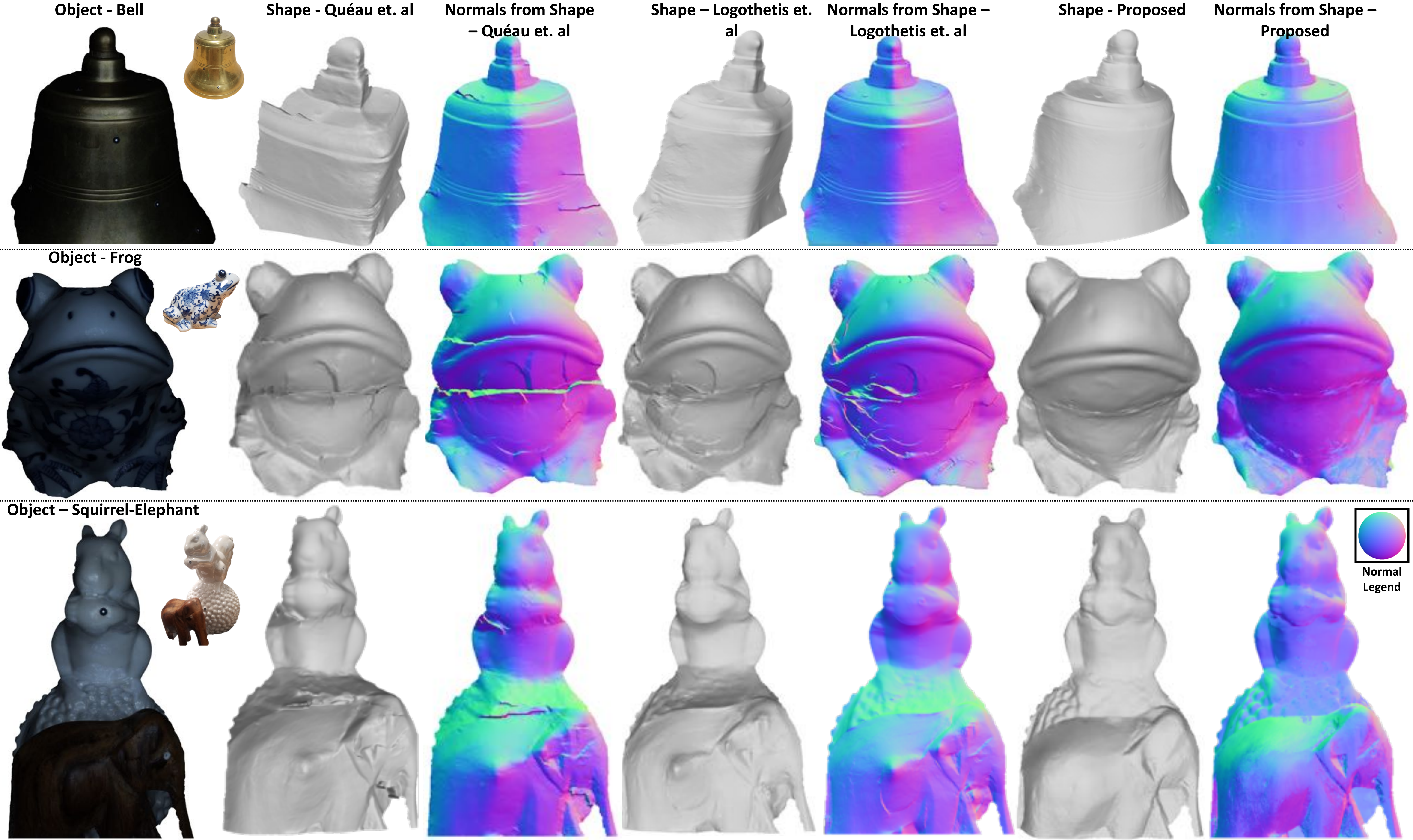}
\vspace{-0.25cm}
\caption{Qualitative comparison of the proposed method to Qu{\'e}au et. al~\cite{queau2018led} and Logothetis et. al~\cite{logothetis2017semi}. The first column shows the average Photometric Stereo image. In contrast to competition, the proposed approach has no visible deformation on the metallic object or the specular highlight in the middle of the elephant. In addition, there is a smooth recovery of belly of the Frog despite the shadows, as well as the bottom of Squirrel despite self reflection.}
\label{fig:main_real_qualitative}
\end{figure}

\noindent
\textbf{Near field competitors.} Then, we present comparison against the state of the art of NF methods  and \cite{logothetis2017semi} \cite{queau2018led}. They both consist of complex variational optimisations which are assuming simplified reflectance models (mostly diffuse for  \cite{queau2018led} using a robust Cauchy estimator) and Blinn-Phong specular for \cite{logothetis2017semi} using an $\ell _1$ loss function. As expected, they are both severely outperformed on the Halloween and Armadillo objects and compare favorably on the dielectric specular Queen. In fact, \cite{logothetis2017semi} achieve the minimum MAE for the Queen which is not surprising as this material is mostly consistent with the assumed Blinn-Phong model.

\noindent
\textbf{Real Data evaluation.} Finally, we present qualitative comparison with \cite{logothetis2017semi} and \cite{queau2018led} on our challenging real data in Figure~\ref{fig:intro} and \ref{fig:main_real_qualitative}. 
As expected, we outperform the competition and our reconstructions do not exhibit any visible deformation in specular highlights (middle of bell and elephant) or cast shadow regions (belly of frog and bottom of squirrel). The superiority comes from the fact that the network was trained using realistic BRDF samples (Disney and MERL) and a data augmentation strategy was also used to account for shadows and self reflections.


%% file: sections/Conclusion.tex
\vspace{-0.02cm}
\section{Conclusion}
\label{sec:conclusion}
\vspace{-0.01cm}
In this work we presented the first CNN approach tackling the near-field Photometric Stereo problem. We leveraged the capability of CNNs to learn to  predict surface normals from reflectance samples for a wide variety of materials. In order to deal with global effect such as shadows and interreflection, we used an augmentation strategy which allowed our method to handle these physical effects. 
Finally, to address near-field effects such as radial light propagation of point light sources, light attenuation and viewing dependency of the light reflection, we iteratively integrate the normal map so to let the problem depend on the depth.  
The main limitation of the proposed approach is the lack of any explicit surface optimisation hence the accuracy is limited by the network predictions.
An interesting future work would be to avoid the integration step and let the network directly predict surface depth. Additionally, a future direction of research would be to relax the assumption of training on specific light positions and employ weaker light calibration setting.

%% file: bmvc_review.bbl
\begin{thebibliography}{34}
\providecommand{\natexlab}[1]{#1}
\providecommand{\url}[1]{\texttt{#1}}
\expandafter\ifx\csname urlstyle\endcsname\relax
  \providecommand{\doi}[1]{doi: #1}\else
  \providecommand{\doi}{doi: \begingroup \urlstyle{rm}\Url}\fi

\bibitem[Ackermann and Goesele(2015)]{AckermannG15}
J.~Ackermann and M.~Goesele.
\newblock A survey of photometric stereo techniques.
\newblock \emph{Foundations and Trends in Computer Graphics and Vision}, 2015.

\bibitem[Blinn(1977)]{Blinn:1977}
J.~F. Blinn.
\newblock Models of light reflection for computer synthesized pictures.
\newblock In \emph{SIGGRAPH}, 1977.

\bibitem[Burley(2012)]{burley2012physically}
B.~Burley.
\newblock Physically-based shading at disney.
\newblock In \emph{SIGGRAPH Course Notes}, 2012.

\bibitem[Chen et~al.(2018)Chen, Han, and Wong]{ChenHW18}
G.~Chen, K.~Han, and K.{-}Y.~K. Wong.
\newblock {PS-FCN:} {A} flexible learning framework for photometric stereo.
\newblock In \emph{ECCV}, 2018.

\bibitem[Chen et~al.(2019)Chen, Han, Shi, Matsushita, and Wong]{chen2019self}
G.~Chen, K.~Han, B.~Shi, Y.~Matsushita, and K.-Y.~K. Wong.
\newblock Self-calibrating deep photometric stereo networks.
\newblock In \emph{CVPR}, 2019.

\bibitem[Clark(1992)]{Clark1992}
J.~J. Clark.
\newblock Active photometric stereo.
\newblock In \emph{CVPR}, 1992.

\bibitem[Hinton(2009)]{Hinton09}
G.~E. Hinton.
\newblock Deep belief networks.
\newblock \emph{Scholarpedia}, 2009.

\bibitem[Ikehata(2018)]{ikehata2018cnn}
S.~Ikehata.
\newblock Cnn-ps: Cnn-based photometric stereo for general non-convex surfaces.
\newblock In \emph{ECCV}, 2018.

\bibitem[Ikehata and Aizawa(2014)]{6909677}
S.~Ikehata and K.~Aizawa.
\newblock Photometric stereo using constrained bivariate regression for general
  isotropic surfaces.
\newblock In \emph{CVPR}, 2014.

\bibitem[Ikehata et~al.(2012)Ikehata, Wipf, Matsushita, and
  Aizawa]{Ikehata2012Robust}
S.~Ikehata, D.~Wipf, Y.~Matsushita, and K.~Aizawa.
\newblock {Robust photometric stereo using sparse regression}.
\newblock In \emph{CVPR}, 2012.

\bibitem[Iwahori et~al.(1990)Iwahori, Sugie, and Ishii]{Iwahori1990point}
Y.~Iwahori, H.~Sugie, and N.~Ishii.
\newblock {Reconstructing shape from shading images under point light source
  illumination}.
\newblock In \emph{ICPR}, 1990.

\bibitem[Ju et~al.(2018)Ju, Qi, Zhou, Dong, and Lu]{JuQZDL18}
Y.~Ju, L.~Qi, H.~Zhou, J.~Dong, and L.~Lu.
\newblock Demultiplexing colored images for multispectral photometric stereo
  via deep neural networks.
\newblock \emph{{IEEE} Access}, 2018.

\bibitem[Lee and Brady(1991)]{Lee1991glaucoma}
S.~Lee and M.~Brady.
\newblock Integrating stereo and photometric stereo to monitor the development
  of glaucoma.
\newblock \emph{Image and Vision Computing}, 1991.

\bibitem[Liu et~al.(2018)Liu, Narasimhan, and Dubrawski]{liu2018near}
C.~Liu, S.~G Narasimhan, and A.~W. Dubrawski.
\newblock Near-light photometric stereo using circularly placed point light
  sources.
\newblock In \emph{ICCV}, 2018.

\bibitem[Logothetis et~al.(2016)Logothetis, Mecca, Qu{\'e}au, and
  Cipolla]{logothetis2016near}
F.~Logothetis, R.~Mecca, Y.~Qu{\'e}au, and R.~Cipolla.
\newblock Near-field photometric stereo in ambient light.
\newblock In \emph{BMVC}, 2016.

\bibitem[Logothetis et~al.(2017)Logothetis, Mecca, and
  Cipolla]{logothetis2017semi}
F.~Logothetis, R~Mecca, and R.~Cipolla.
\newblock Semi-calibrated near field photometric stereo.
\newblock In \emph{CVPR}, 2017.

\bibitem[Logothetis et~al.(2019)Logothetis, Mecca, and
  Cipolla]{logothetis2019differential}
F.~Logothetis, R.~Mecca, and R.~Cipolla.
\newblock A differential volumetric approach to multi-view photometric stereo.
\newblock In \emph{ICCV}, 2019.

\bibitem[Logothetis et~al.(2020)Logothetis, Budvytis, Mecca, and
  Cipolla]{logothetis2020pxnet}
F.~Logothetis, I.~Budvytis, R.~Mecca, and R.~Cipolla.
\newblock {PX-NET}: {Simple, Efficient Pixel-Wise Training of Photometric
  Stereo Networks}.
\newblock In \emph{arXiv preprint {arXiv}:2008.04933}, 2020.

\bibitem[Matusik et~al.(2003)Matusik, Pfister, Brand, and McMillan]{Matusik03}
W.~Matusik, H.~Pfister, M.~Brand, and L.~McMillan.
\newblock A data-driven reflectance model.
\newblock \emph{ACM Transactions on Graphics}, 2003.

\bibitem[Mecca et~al.(2014)Mecca, Wetzler, Bruckstein, and
  Kimmel]{Mecca2014near}
R.~Mecca, A.~Wetzler, A.~Bruckstein, and R.~Kimmel.
\newblock {Near Field Photometric Stereo with Point Light Sources}.
\newblock \emph{SIAM Journal on Imaging Sciences}, 2014.

\bibitem[Mecca et~al.(2016)Mecca, Qu\'eau, Logothetis, and
  Cipolla]{MeccaQLC2016}
R.~Mecca, Y.~Qu\'eau, F.~Logothetis, and R.~Cipolla.
\newblock A single lobe photometric stereo approach for heterogeneous material.
\newblock \emph{SIAM Journal on Imaging Sciences}, 2016.

\bibitem[Onn and Bruckstein(1990)]{Onn1990}
R.~Onn and A.~Bruckstein.
\newblock {Integrability disambiguates surface recovery in two-image
  photometric stereo}.
\newblock \emph{IJCV}, 1990.

\bibitem[Prados and Faugeras(2003)]{Prados2003}
E.~Prados and O.~Faugeras.
\newblock Perspective shape from shading and viscosity solutions.
\newblock In \emph{ICCV}, 2003.

\bibitem[Qu\'eau and Durou(2015)]{queau2015edge}
Y.~Qu\'eau and J.-D. Durou.
\newblock Edge-preserving integration of a normal field: Weighted least
  squares, {TV} and {L1} approaches.
\newblock In \emph{SSVM}, 2015.

\bibitem[Qu{\'e}au et~al.(2018)Qu{\'e}au, Durix, Wu, Cremers, Lauze, and
  Durou]{queau2018led}
Y.~Qu{\'e}au, B.~Durix, Tao Wu, D.~Cremers, F.~Lauze, and J.-D. Durou.
\newblock Led-based photometric stereo: Modeling, calibration and numerical
  solution.
\newblock \emph{JMIV}, 2018.

\bibitem[Santo et~al.(2017)Santo, Samejima, Sugano, Shi, and
  Matsushita]{santodeep}
H.~Santo, M.~Samejima, Y.~Sugano, B.~Shi, and Y.~Matsushita.
\newblock Deep photometric stereo network.
\newblock In \emph{ICCV Workshops}, 2017.

\bibitem[Smith and F.(2016)]{SmithFang2016}
{W.A.P.} Smith and Fufu F.
\newblock Height from photometric ratio with model-based light source
  selection.
\newblock \emph{CVIU}, 2016.

\bibitem[Tang et~al.(2012)Tang, Salakhutdinov, and Hinton]{TangSH12a}
Y.~Tang, R.~Salakhutdinov, and G.~E. Hinton.
\newblock Deep lambertian networks.
\newblock In \emph{ICML}, 2012.

\bibitem[Taniai and Maehara(2018)]{taniai2018neural}
T.~Taniai and T.~Maehara.
\newblock Neural inverse rendering for general reflectance photometric stereo.
\newblock In \emph{ICML}, 2018.

\bibitem[Tankus and Kiryati(2005)]{Tankus2005}
A.~Tankus and N.~Kiryati.
\newblock Photometric stereo under perspective projection.
\newblock In \emph{ICCV}, 2005.

\bibitem[Wetzler et~al.(2014)Wetzler, Mecca, Bruckstein, and Kimmel]{WMBK14}
A.~Wetzler, R.~Mecca, A.~M. Bruckstein, and R.~Kimmel.
\newblock Close-range photometric stereo with point light sources.
\newblock In \emph{3DV}, 2014.

\bibitem[Woodham(1980)]{Woodham1980}
R.~J. Woodham.
\newblock Photometric method for determining surface orientation from multiple
  images.
\newblock \emph{Optical Engineering}, 1980.

\bibitem[Yu and Smith(2017)]{YuS17}
Y.~Yu and W.~A.~P. Smith.
\newblock Pvnn: A neural network library for photometric vision.
\newblock In \emph{ICCV Workshop}, 2017.

\bibitem[Yuille et~al.(1999)Yuille, Snow, Epstein, and Belhumeur]{YuilleSEB99}
A.~L. Yuille, D.~Snow, R.~Epstein, and P.~N. Belhumeur.
\newblock Determining generative models of objects under varying illumination:
  Shape and albedo from multiple images using {SVD} and integrability.
\newblock \emph{IJCV}, 1999.

\end{thebibliography}
